\documentclass[pdflatex,sn-mathphys-num]{sn-jnl}

\usepackage[T1]{fontenc}
\usepackage{graphicx}%
\usepackage{multirow}%
\usepackage{multicol}
\usepackage{amsmath,amssymb,amsfonts}%
\usepackage{amsthm}%
\usepackage{mathrsfs}%
\usepackage[title]{appendix}%
\usepackage{textcomp}%
\usepackage{manyfoot}%
\usepackage{booktabs}%
\usepackage{algorithm}%
\usepackage{algorithmicx}%
\usepackage{algpseudocode}%
\usepackage{listings}%
\usepackage{hyperref}       
\usepackage{nicefrac}       
\usepackage{microtype}      
\usepackage{centernot}
\usepackage{subcaption}
\usepackage{adjustbox}
\usepackage{array}
\usepackage{hhline} 
\usepackage{nicematrix} 

\usepackage{soul}
\usepackage{xcolor,colortbl} 

\usepackage{url}            

\usepackage{draftwatermark}
\SetWatermarkText{Under-review}
\SetWatermarkScale{0.5} 

\theoremstyle{thmstyleone}%
\theoremstyle{thmstyletwo}%

\theoremstyle{thmstylethree}%

\begin{document}

\title[Article Title]{KGroups: A Versatile Univariate Max-Relevance Min-Redundancy Feature Selection Algorithm for High-dimensional Biological Data}


\author*[1,3]{\fnm{Malick} \sur{Ebiele}}\email{malick.ebiele@adaptcentre.ie}

\author[2,3]{\fnm{Malika} \sur{Bendechache}}\email{malika.bendechache@universityofgalway.ie }

\author[1,3]{\fnm{Rob} \sur{Brennan}}\email{rob.brennan@ucd.ie}

\affil*[1]{\orgdiv{Computer Science}, \orgname{University College Dublin}, \orgaddress{ \city{Dublin}, \state{Dublin}, \country{Ireland}}}

\affil[2]{\orgdiv{Computer Science}, \orgname{University of Galway}, \orgaddress{ \city{Galway}, \state{Galway}, \country{Ireland}}}

\affil[3]{\orgname{ADAPT Centre}, \orgaddress{\city{Dublin}, \state{Dublin}, \country{Ireland}}}


\abstract{
This paper proposes a new univariate filter feature selection (FFS) algorithm called KGroups. The majority of work in the literature focuses on investigating the relevance or redundancy estimations of feature selection (FS) methods. This has shown promising results and a real improvement of FFS methods' predictive performance. However, limited efforts have been made to investigate alternative FFS algorithms. This raises the following question: how much of the FFS methods' predictive performance depends on the selection algorithm rather than the relevance or the redundancy estimations? The majority of FFS methods fall into two categories: relevance maximisation (Max-Rel, also known as KBest) or simultaneous relevance maximisation and redundancy minimisation (mRMR). KBest is a univariate FFS algorithm that employs sorting (descending) for selection. mRMR is a multivariate FFS algorithm that employs an incremental search algorithm for selection. In this paper, we propose a new univariate mRMR called KGroups that employs clustering for selection. Extensive experiments on 14 high-dimensional biological benchmark datasets showed that KGroups achieves similar predictive performance compared to multivariate mRMR while being up to 821 times faster. KGroups is parameterisable, which leaves room for further predictive performance improvement through hyperparameter finetuning, unlike mRMR and KBest. KGroups outperforms KBest.
}




\maketitle

\section{Introduction}\label{sec:Introduction}

Feature selection (FS) methods in machine learning (ML) and statistical analysis have existed since at least the 1990s with an ever increasing interest \cite{theng_feature_2024}. They define ways of selecting a subset of features from the original dataset while maintaining a competitive predictive performance. Feature selection methods are important in tackling computational cost (in terms of training time) and resource requirements \cite{guyon_introduction_2003}. Today, they are also important for addressing the energy consumption, efficiency, monetary cost of training, and the sustainability of Artificial Intelligence (AI) models \cite{verdecchia_data-centric_2022,omar_energy_2024,cottier_rising_2024}. 

There are three primary type of FS methods: filters, wrappers, and embedded. Wrappers are considered the most predictively performant, while filters are considered the most computationally efficient. Filters are also considered the most generalisable to other models (models they were not initially trained for; see Table \ref{tab:primary_feature_selection_types} in section \ref{sec:Background} below) \cite{guyon_introduction_2003, jiao_survey_2023, theng_feature_2024}. 

With the rise of interest in green AI initiatives, filter feature selection (FFS) methods are receiving more attention from researchers and practitioners because they align with sustainable AI objectives \cite{pedrycz_towards_2022,tabbakh_towards_2024,bolon-canedo_review_2024,salehi_data-centric_2024,jarvenpaa_synthesis_2024,wu_sustainable_2022}. This can also be explained by the fact that FFS methods have only one drawback to address (lag in predictive performance) compared to two drawbacks for wrappers (computational inefficiency and failure to generalise to other models). FFS methods have been widely applied to feature selection and dimensionality reduction tasks in heterogeneous domains including but not limited to biology, healthcare, marketing, information retrieval, human activity recognition \cite{bus_using_2022,doewes_feature_2017,dubey_cosine-similarity_2017,zhao_maximum_2019,atef_novel_2025,billah_minimum_2020,radovic_minimum_2017,ma_wavelength_2024,shirzad_feature_2015,billah_minimum_2020,huang_multilabel_2020}.

FS methods can be classified into two groups: value-based and non-value-based. Value-based FS methods first assign a relevance or a redundancy value to each feature then perform the selection using either or both of the assigned relevance and redundancy values. Max-Rel, mRMR, and KGroups are all value-based FFS. Non-value-based FS techniques perform the selection task without assigning any values to the features; they are in majority wrappers (see Table \ref{tab:primary_feature_selection_types} in the background section). This paper focuses on value-based FFS methods because they are computationally efficient, have only one drawback, and can be integrated with explainable AI \cite{cheng_comprehensive_2024}. Value-based FFS estimate features relevance using a similarity measure from i) information theory (e.g. mutual information also known as information gain), ii) statistics (e.g. F-value also known as F-statistics), or iii) machine learning models' built-in feature ranking methods also known as embedded FS techniques. Once a relevance value is assigned to each of the feature, the features are sorted in descending order of relevance and the top $k$ are selected (thus the names Max-Rel, SelectKBest or KBest). As to multivariate FFS methods, a redundancy estimator is needed in addition to the relevance estimator. They consist of maximising the difference between or the quotient of the relevance $\text{Rel}(F_i)$ by the redundancy $\text{Red} (F_i; \mathcal{S})$ of the candidate feature $F_i$ with respect to the already selected features $\mathcal{S}$ using an incremental search algorithm (see section \ref{sec:Related_Work}, particularly Eqs. (\ref{eq:mRMR_general_formula}) and (\ref{eq:mRMR_quotient_formula})). These methods are referred to as mRMR meaning Min-Redundancy Max-Relevance.

This paper proposed a new versatile univariate algorithm, called KGroups, that simultaneously maximises the relevance and minimises the redundancy using clustering. In other words, KGroups is a univariate mRMR. First, a relevance value is assigned to each feature. Then, the features are clustered into $k$ groups using binning. Finally, the feature with the maximum relevance value from each cluster is selected (for more details, see section \ref{sec:Proposed_Method} below). There are been previous attempts to use binning-based clustering or discretisation for FS and dimensionality reduction \cite{dubey_cosine-similarity_2017,liu_spatial_2013,muthukumaran_novel_2016, sanders_adaptive_2001, zhang_compact_2014}. KGroups extends the Supervised Cosine Similarity Information Gain (SCSIG) proposed by \citet{dubey_cosine-similarity_2017}. SCSIG was originally designed for cosine similarity-based binning and mutual information-based selection (see section \ref{sec:Proposed_Method} for more details).

The paper addresses the following research question:
\textbf{To what extent can a univariate Max-Relevance Min-Redundancy filter feature selection algorithm improve the performance of feature selection methods in terms of classification accuracy and computational cost?} To answer the research question above, we performed extensive experimental analysis of KGroups with competing algorithms (KBest, and multivariate mRMR) on 14 high-dimensional biological benchmark datasets and reported their respective average 5-fold cross validation accuracy score and execution time (see section \ref{sec:Experimental_Results}). \textbf{This paper has the following contributions:} (1) Design and implementation of KGroups, a versatile univariate mRMR algorithm that competes with the state of the art multivariate mRMR FFS algorithm while being up to 821 times faster. (2) Extension of the SCSIG \cite{dubey_cosine-similarity_2017} FFS method into a flexible algorithm that support any type of relevance estimator. KGroups is parameterisable and introduces the concept of tie-breaking; unlike SCSIG. (3) Comprehensive Analysis of KGroups, highlighting its strengths, weaknesses, and potentials.

The remainder of the paper is as follows. The next section defines the cores concepts of FFS methods. In section \ref{sec:Related_Work}, we present the related work. Section \ref{sec:Proposed_Method} and \ref{sec:Experimental_Setting} describes our proposed method and the experimental setting, respectively. The experimental results and discussion are presented in section \ref{sec:Experimental_Results}. The limitations and potentials of the KGroups algorithm is presented in section \ref{sec:KGroups Limitations and Potentials}. Finally, the concluding remarks and future work is highlighted in section \ref{sec:Conclusion}.

\section{Background}\label{sec:Background}

In this section, we provide the definitions of the core concepts of the Feature Selection (FS) domain. 
\subsection{Feature Selection Methods}
There are three primary types of FS methods: filters, wrappers, and embedded \cite{guyon_introduction_2003, jiao_survey_2023, moslemi_tutorial-based_2023,theng_feature_2024}.
Table \ref{tab:primary_feature_selection_types} below gives a comparative analysis of the three primary types of FS in terms of relative predictive performance, computational cost, and generalisation [of the selected features] to other models. It also provides a brief description of each method and examples.

\begin{table}[!htp]
\centering
\small
\caption{Comparative analysis of the primary types of feature selection methods (adapted from \cite{jiao_survey_2023})}
\label{tab:primary_feature_selection_types}
\begin{tabular}{p{0.085\textwidth}p{0.075\textwidth}p{0.075\textwidth}p{0.09\textwidth}p{0.15\textwidth}p{0.4\textwidth}}
\toprule
\textbf{Type} & \textbf{Predic-tive Perfor-mance} & \textbf{Compu-tational Cost} & \textbf{Genera-lisation to other models}& \textbf{Example Methods}& \textbf{Short Description}\\
\midrule
Filters & Low & Low & High & 
SelectKBest; SelectPercentile &
\begin{minipage}[t]{\linewidth}
- Perform feature selection at pre-processing without relying on ML model \\
- Use statistics, information theory, or geometry to measure relevance and redundancy \\
- Model-agnostic approach
\end{minipage} \\
\hline
Wrappers & High & High & Low &
Forward Selection; Backward Selection; Recursive Selection; Exhaustive Selection &
\begin{minipage}[t]{\linewidth}
- Integrate variable selection into training process \\
- Evaluate subsets based on predictive performance \\
- Can use SHAP values, permutation importance \\
- Assigned values can be negative
\end{minipage} \\
\hline
Embedded & Medium & Medium & Medium &
Gini and Information Gain Importance &
\begin{minipage}[t]{\linewidth}
- Use ML models with built-in feature selection \\
- Tied to specific models (e.g. tree-based models) \\
- Assign importance values during training \\
- Non-relevant features often assigned zero
\end{minipage} \\
\bottomrule
\end{tabular}
\end{table}

\subsection{Filter Feature Selection Methods}
According to Table \ref{tab:primary_feature_selection_types}, FFS methods have only one shortcoming which is their low predictive performance compared to wrapper and embedded methods. In this paper, we propose a value-based FFS algorithm that achieves similar average 5-fold cross validation accuracy compared to competing algorithms across 14 benchmark datasets (see Tables \ref{tab:Best overall configs - KGvsKB} and \ref{tab:Best overall configs - KGvsMRMR}), three heterogenous relevance estimators (see Eqs. (\ref{eq:mutual_info})-(\ref{eq:gini_index}) below), and 6 popular classifiers (see section \ref{subsec:classifiers} for more details about the classification models).

\label{subsec:scoring_functions} 
For relevance estimation, this paper employed mutual information, also known as information gain ($I$), F-value aka F-statistics ($F$), and Random Forest Gini importance (Gini), also known as Mean Decrease in Impurity (MDI); see Eqs. (\ref{eq:mutual_info}), (\ref{eq:Fvalue}), and (\ref{eq:gini_index}), respectively.

\begin{equation}\label{eq:mutual_info}
     I(X, Y) = I(Y, X)= \displaystyle\sum_{x_i\in X}\sum_{y_i\in Y} P(x_i, y_i) \log\Big( \frac{P(x_i, y_i)}{P(x_i)P(y_i)}\Big)
\end{equation}

\begin{equation}\label{eq:Fvalue}
    \text{F-value} = \frac{\text{Explained variance}}{\text{Unexplained variance}} = \frac{\text{Between-sample variability}} {\text{Within-sample variability}} \text{, } 
\end{equation}
The $F$ calculation involves many steps which are too long to present here. The formula above (Eq. (\ref{eq:Fvalue})) is just illustrative. A comprehensive and detailed explanation can be found in \citet{siegel_chapter_2016}. 

\begin{equation}\label{eq:gini_index}
     \text{Gini}_{x_j} = \mathrm{MDI}_{x_j} = \frac{1}{T} \sum_{t=1}^{T} \sum_{\substack{\text{nodes } n \\ \text{split on } x_j}} p(n) \left[ I(n) - p_{\text{left}} I(\text{left}) - p_{\text{right}} I(\text{right}) \right]
\end{equation}
In the Gini importance formula (Eq. (\ref{eq:gini_index})) , $T$ is the number of trees in the forest; $n$ the node in tree $t$ that splits on feature $j$; $p(n)$ the proportion of samples reaching node $n$; $I(n)$ the impurity of node $n$ (in terms of the Gini index); $p_{\mathit{left}}$ and $p_{\mathit{right}}$ are the proportions of samples going to the left and right child, respectively; and $I(\text{left}),\ I(\text{right})$ the impurity of the left and right children, respectively. The Gini index is defined as $G = 1 - \displaystyle \sum_{c_j \in \{Y\}} p(c_j)^2$, with $\{Y\}$ the set of unique classes in $Y$ and $p(c_j)$ the probability of samples belonging to class $c_j$ in the node or dataset.

\section{Related Work}\label{sec:Related_Work}
In this section, we review the existing FFS literature. Before doing so, we provide an overview of the FFS methods as found in literature in Fig. \ref{fig:FS steps}. This figure illustrates the key steps and decision points involved in current FFS approaches. Section \ref{subsec:General formulation of Max-Rel and mRMR} below gives the general formulation of the state of the art FFS algorithms (Max-Rel and mRMR). The vast majority of the FFS methods found in the literature is an instance of one of these general formulation shown in Eqs. (\ref{eq:MaxRel_general_formula})-(\ref{eq:mRMR_quotient_formula}); independently from their respective relevance and redundancy estimators they use.

\subsection{General formulation of Max-Rel and mRMR}\label{subsec:General formulation of Max-Rel and mRMR}
The equations (\ref{eq:MaxRel_general_formula}) and (\ref{eq:mRMR_general_formula}) are the general formulation of the Max-Rel and mRMR feature selection problems. $\mathcal{F}$ is the set of all features, $\mathcal{S}$ is the set of selected features, $\mathcal{F\setminus S}$ is the set of non-selected features, and $F_i$ is the candidate feature. $\text{Rel}(F_i)$ is the relevance value of $F_i$ and $\text{Red}(F_i, \mathcal{S})$ is the redundancy value of $F_i$ with respect to the already selected features. 

The objective of Max-Rel is to add $F_i$ to $\mathcal{S}$ if and only if $\text{Rel}(F_i)$ is the maximum relevance value of all the features in $\mathcal{F\setminus S}$. To do so, Max-Rel uses a descending sorting algorithm and return the top $k$ of the features with $k\leq |\mathcal{F}|$.

mRMR, on the other hand, aims to add $F_i$ to $\mathcal{S}$ if and only if $F_i$ maximises the difference or quotient between its relevance $\text{Rel}(F_i)$ and its redundancy $\text{Red}(F_i, \mathcal{S})$ with respect to the already selected features. The first formulation by \citet{battiti_using_1994} includes a redundancy regulation parameter $\beta \in [0,1]$. This  parameter $\beta$ is set to 1 in the subsequent studies, starting from \citet{ding_minimum_2005}. $\beta=1$ is the standard in recent mRMR-based FFS methods (Eq. (\ref{eq:mRMR_subtraction_formula})). \citet{ding_minimum_2005} proposed another reformulation of mRMR by replacing the subtraction operation between relevance and redundancy by a division operation (see Eq. (\ref{eq:mRMR_quotient_formula})). More details on mRMR-based methods are provided in the section \ref{subsec:Multivariate FFS}.

\begin{align}
&\max_{F_i \in \mathcal{F\setminus S}} \text{Rel}(F_i)
\label{eq:MaxRel_general_formula}\\
&\max_{F_i \in \mathcal{F\setminus S} } \text{Rel}(F_i) - \beta\times \text{Red}(F_i; \mathcal{S})
\label{eq:mRMR_general_formula}\\
&\max_{F_i \in \mathcal{F\setminus S} } \text{Rel}(F_i) - \text{Red}(F_i; \mathcal{S}) 
\label{eq:mRMR_subtraction_formula}\\
&\max_{F_i \in \mathcal{F\setminus S} } \frac{\text{Rel}(F_i)}{\text{Red}(F_i; \mathcal{S})}  
\label{eq:mRMR_quotient_formula}
\end{align}

FFS methods can be classified into two main groups: \textbf{univariate and multivariate}. This classification mainly focus on the number of variable involved in the actual selection algorithm i.e. \textbf{sorting or an incremental search algorithm} i.e. sequential forward search. For instance, a feature selection method using random forest Gini importance (an embedded FS method) can be classified as univariate FFS even though the relevance estimation involves many features and their interactions \cite{guyon_introduction_2003}. On the other hand, multivariate FFS incorporate an estimation of feature interaction (redundancy or complementarity) and uses a search algorithm such as the sequential forward search (a greedy search algorithm) to select the optimal feature subset \cite{li_new_2020,battiti_using_1994,ding_minimum_2005,ihianle_minimising_2024}. Figure \ref{fig:FS steps} below give an overview of FFS methods on a tabular data as found in the literature.

\begin{figure}[!h]
\centering
\includegraphics[width=0.9\textwidth]{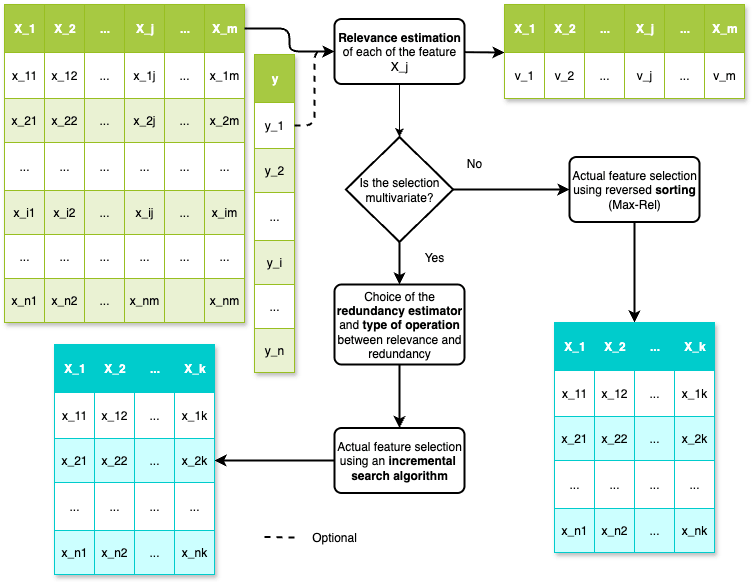}
\caption{Overview of FFS methods on a tabular data as found in the literature.}\label{fig:FS steps}
\end{figure}

\subsection{Univariate FFS}\label{subsec:Univariate FFS}
This is the most used family of FFS methods because they are the most computational efficient and easily adaptable from one relevance estimator to another \cite{bommert_benchmark_2022, bugata_aspects_2019,omar_energy_2024}. Descending sorting is the only search algorithm used for Max-Rel feature selection methods. 

Even though they are considered less predictively performant than multivariate FFS methods (mRMR), variance-based Max-Rel has been shown to outperform mRMR and embedded FS methods (including random forest Gini importance, Permutation importance, and Boosting importance from the eXtreme Gradient Boosting model) in a systematic benchmark study on high-dimensional biological datasets \cite{bommert_benchmark_2022}. This kind of studies are missing in the extant literature, especially in recent years. These results are not surprising but highlight the challenges of the FS task where every single decision can positively or negatively impact the expected outcome. FS challenges and gaps are discussed in section \ref{subsec:Challenges of the FS task} below.  

\subsection{Multivariate FFS}\label{subsec:Multivariate FFS}
The first Min-Redundancy Max-Relevance (mRMR) FFS method was introduced by \citet{battiti_using_1994} and called \textbf{mutual information-based feature selection (MIFS)}. It uses mutual information for both relevance and redundancy estimation (see Eq. (\ref{eq:mifs_formula})). \citet{battiti_using_1994} used an incremental search algorithm (a greedy algorithm) to solve the objective function shown in Eq. (\ref{eq:mifs_formula}). 

\begin{align}
&\max_{X_i \in \mathcal{F\setminus S} } I(X_i; Y) - \beta\times \sum_{X_j \in \mathcal{S}} I(X_i; X_j)
\label{eq:mifs_formula},
\end{align}
where $\text{Rel}(X_i)= I(X_i; Y)$ and $\text{Red} (X_i; \mathcal{S}) = \sum_{X_j \in \mathcal{S}} I(X_i; X_j)$.

Recent mRMR-based FFS methods used one of the two formulation proposed by \citet{ding_minimum_2005} shown in Eqs. (\ref{eq:mRMR_subtraction_formula}) and (\ref{eq:mRMR_quotient_formula}). They also proposed a modified version of the mutual information-based redundancy estimation such as $\text{Red} (X_i; \mathcal{S}) = \frac{1}{|\mathcal{S}|} \sum_{X_j \in \mathcal{S}} I(X_i; X_j)$. \citet{ding_minimum_2005} also proposed a new relevance and redundancy estimator using F-value and Pearson correlation coefficient, respectively. They also used an incremental search algorithm to solve the objective functions of their proposed mRMR-based FFS methods. 
The four variants proposed by \citet{ding_minimum_2005} are shown in Eqs. (\ref{eq:MID_formula})-(\ref{eq:FCQ_formula}).
\begin{align}
&\max_{X_i \in \mathcal{F\setminus S} }  I(X_i; Y) -  \frac{1}{|\mathcal{S}|} \sum_{X_j \in \mathcal{S}} I(X_i; X_j)
\label{eq:MID_formula}\\
&\max_{X_i \in \mathcal{F\setminus S} } \frac{I(X_i; Y)}{\frac{1}{|\mathcal{S}|} \sum_{X_j \in \mathcal{S}} I(X_i; X_j)}  
\label{eq:MIQ_formula}\\
&\max_{X_i \in \mathcal{F\setminus S} }  \text{F-value}(X_i; Y) -  \frac{1}{|\mathcal{S}|} \sum_{X_j \in \mathcal{S}} \rho(X_i; X_j)
\label{eq:FCD_formula}\\
&\max_{X_i \in \mathcal{F\setminus S} } \frac{\text{F-value}(X_i; Y)}{\frac{1}{|\mathcal{S}|} \sum_{X_j \in \mathcal{S}} \rho(X_i; X_j)}  
\label{eq:FCQ_formula}
\end{align}

$\rho(X_i; X_j) = \frac{\operatorname{cov}(X_i; X_j)}{\sigma_{X_i} \, \sigma_{X_j}}$ is Pearson correlation coefficient; where $\operatorname{cov}(X_i; X_j)$ is the covariance of $X_i$ and $X_j$; $\sigma_{X_i}$ and $\sigma_{X_j}$ are the standard deviations of  $X_i$ and $X_j$, respectively. 

In the extant literature, most efforts are made on finding alternative relevance or redundancy estimators \cite{jo_improved_2019,che_maximum_2017,radovic_minimum_2017,senawi_new_2017,mandal_improved_2013,yang_minimal-redundancy-maximal-relevance_2013,lin_conditional_2006,kwak_input_2002}. For instance, in a recent study \cite{yuan_feature_2025}, the authors proposed a relevance estimator using the maximum information coefficient (MIC) and a redundancy estimator using the wavelet similarity,

\begin{align}
&  \max_{X_i \in \mathcal{F\setminus S} } MIC(X_i; Y) - \frac{1}{|\mathcal{S}|} \sum_{X_j \in \mathcal{S}} WTSim(X_j; X_i)
\label{eq:WSMI_formula},
\end{align}
where 
\begin{align}
&\text{Rel}(X_i) = MIC(X_i; Y) = \max_{|X_i|, |Y| < B} \frac{I(X_i; Y)}{\log_2 \left( \min(|X_i|, |Y|) \right)},\\
&\text{Red}(X_i; \mathcal{S}) = \frac{1}{|\mathcal{S}|} \sum_{X_j \in \mathcal{S}} WTSim(X_j; X_i) = \frac{1}{|\mathcal{S}|} \sum_{X_j \in \mathcal{S}}\overline{Similarity(hash(X_j), hash(X_i))},
\label{eq:MIC_WTSim_formulae}
\end{align}
$B=|\mathcal{F}|^{0.6}$, and $\overline{Similarity(hash(X_j), hash(X_i))}$ is the similarity of the features at different scales after wavelet decomposition. For more details, please refer to \citet{yuan_feature_2025}. 

Another example is found in \citet{yuan_cscim_fs_2023} where the objective function is defined as 
\begin{align}
& \max_{X_i \in \mathcal{F\setminus S} }  SU(X_i ; Y) -  \frac{1}{|\mathcal{S}|} \sum_{x_j \in \mathcal{S}} Sim_{normal}(hash(X_j), hash(X_i)),
\label{eq:SU_CosTansSim_formulae}
\end{align}
where 
\begin{align}
&\text{Rel}(X_i) = SU(X_i ; Y) = \frac{2 \cdot I(X_i ; Y)}{H(X_i) + H(Y)}, \label{eq:SU_Rel_formula} \text{ and}\\ 
&\text{Red}(X_i; \mathcal{S}) = \frac{1}{|\mathcal{S}|} \sum_{x_j \in \mathcal{S}} Sim_{normal}(hash(X_j), hash(X_i)).
\label{eq:CosTansSim_Red_formula}
\end{align}
$SU$ is the symmetry uncertainty, and $Sim_{normal}(hash(X_j), hash(X_i))$ is the normalised similarity between $X_i$ and $X_j$ after a cosine transform (like a Fourier transform used in signal processing); for more details, please refer to \cite{yuan_cscim_fs_2023}. 

\citet{zhao_maximum_2019} proposed three new extensions of mRMR by introducing random forest Gini importance as relevance estimator and the randomized dependence coefficient (RDC) as a [non-linear] redundancy estimator. The first extension consists of replacing $\rho(X_i; X_j)$ by $\rho_{rdc}(X_i; X_j)$ in Eq. (\ref{eq:FCQ_formula}). The second consists of replacing $\text{F-value}(X_i; Y)$ by random forest Gini importance in Eq. (\ref{eq:FCQ_formula}). The third and last extension consists of substituting $\text{F-value}(X_i; Y)$ by random forest Gini importance and $\rho(X_i; X_j)$ by $\rho_{rdc}(X_i; X_j)$ in Eq. (\ref{eq:FCQ_formula}). The randomized dependence coefficient $\rho_{rdc}(X_i; X_j)$ is defined as the largest canonical correlation between the non-linear projections of $X_i$ and $X_j$; for more details, please refer to \citet{lopez-paz_randomized_2013}. \citet{ihianle_minimising_2024} also proposed four new extensions of mRMR. The first two consists of replacing $\rho(X_i; X_j)$ by $\mathrm{dCor}(X_i; X_j)$ in Eqs. (\ref{eq:FCD_formula}) and (\ref{eq:FCQ_formula}). $\mathrm{dCor}(X_i; X_j) = \frac{\mathrm{dCov}(X_i, X_j)}{\sqrt{\mathrm{dVar}(X_i) \, \mathrm{dVar}(X_j)}}$ is the distance correlation coefficient between $X_i$ and $X_j$. It is a statistical measure that evaluates the interdependence between two random variables by considering both linear and nonlinear relationships \cite{szekely_brownian_2009}. The last two consists of replacing $\rho(X_i; X_j)$ by $MIC(X_i; X_j)$ in Eqs. (\ref{eq:FCD_formula}) and (\ref{eq:FCQ_formula}). \textbf{There are no dedicated relevance or redundancy estimators}; meaning, any relevance estimator can be employed as a redundancy estimator, and vice versa. For instance, \citet{yuan_feature_2025} employed $MIC$ as a relevance estimator while \citet{ihianle_minimising_2024} used it as a redundancy estimator. 

Some studies include conditionality in the estimation of the relevance or redundancy \cite{zhang_conditional-weight_2021,zhang_feature_2020,lin_conditional_2006}. For instance, \citet{zhang_feature_2020} redefined mRMR as shown in Eq. (\ref{eq:UCRFS_formula}) below by including conditional mutual information and conditional entropy terms in the redundancy estimator. In a more recent study \cite{wang_feature_2023}, the authors also defined redundancy in terms of a conditional mutual information as shown in Eq. (\ref{eq:MRMSR_formula}). They redefined the relevance estimator as follow $\text{Rel}(X_i) = \frac{I(X_i; Y)}{H(Y)}$. Given that both, the relevance and redundancy estimators, have the same denominator, by multiplying Eq. (\ref{eq:MRMSR_formula}) by $H(Y)$ one gets Eq. (\ref{eq:MRMSR_formula_rewritten}). Given the common denominator, their redefinition can be reduced to Eq. (\ref{eq:MRMSR_formula_rewritten}) below. 

\begin{align}
&\max_{X_i \in \mathcal{F\setminus S} }  \left[ I(X_i; Y) - \left( \frac{1}{|\mathcal{S}|} \sum_{X_j \in S} I(X_i; X_j) - \sum_{X_j \in \mathcal{S}} \frac{I(X_j; Y | X_i)}{H(Y|X_i) + H(Y|X_i, X_j)} \right)\right]
\label{eq:UCRFS_formula}
\end{align}

\begin{align}
&\max_{X_i \in \mathcal{F\setminus S} } \left\{ \frac{I(X_i; Y)}{H(Y)} - \left[ \frac{1}{|\mathcal{S}|} \sum_{X_j \in \mathcal{S}} \left( 1 -\frac{I(X_j; Y \mid X_i) + I(X_i; Y \mid X_j)}{2H(Y)} \right) \right] \right\}
\label{eq:MRMSR_formula}
\end{align}

\begin{align}
&\max_{X_i \in \mathcal{F\setminus S} } \left[ I(X_i; Y) - \left( H(Y) -  \frac{1}{2|\mathcal{S}|} \sum_{X_j \in \mathcal{S}} I(X_j; Y \mid X_i) + I(X_i; Y \mid X_j) \right) \right]
\label{eq:MRMSR_formula_rewritten}
\end{align}

All those mRMR-based methods found in the literature still used the original incremental search algorithm algorithm (or a variant) to select subsequent features from $\mathcal{F\setminus S}$. That makes all of them computational expensive, and therefore, they generalised poorly when $|\mathcal{F}|$ increases considerably \cite{ramirez-gallego_fast-mrmr_2017} (also see section \ref{subsec:Execution Time}, particularly Fig. \ref{fig:total_cpu_time} and Table \ref{tab:per_dataset_cpu_time}). Another observation is that none of them investigate both formulae of mRMR (shown in Eqs. (\ref{eq:mRMR_subtraction_formula}) and (\ref{eq:mRMR_quotient_formula})) despite studies showing that it is very difficult to know a priori which one will yield the optimal subset $\mathcal{S}^*$ from a given feature set $\mathcal{F}$ \cite{berrendero_mrmr_2016,bugata_aspects_2019}. 

To address mRMR-based methods computational requirements, \citet{ramirez-gallego_fast-mrmr_2017} proposed two new implementations of mRMR using \textbf{graphics processing units (GPUs)} for parallel computing, and \textbf{Apache Spark} for distributed computing. Another approach could be to employ less computationally expensive relevance and redundancy estimators. For instance, \textbf{mRMR using F-value and Pearson correlation is faster than using mutual information for both relevance and redundancy} because mutual information is more expensive to estimate than F-value and Pearson correlation \cite{zhao_maximum_2019} (also see Fig. \ref{fig:total_cpu_time}). Dynamic programming could be another option to speed up mRMR-based FFS methods. 

\subsection{Challenges of the FS task}\label{subsec:Challenges of the FS task}
FS is a mature domain that has been developing for decades. Despite its maturity, there are still a lot of challenges to address and questions to answer. For instance, given an optimal feature subset $\mathcal{S}^*$, any feature added to or removed from  $\mathcal{S}^*$ can worsen its predictive performance by multiple points. This is because  $\mathcal{S}^*$ predictive performance is determined by individual or non-contextual relevance of each feature as well as interactive or contextual relevance (i.e. redundancy and complementarity) \cite{guyon_introduction_2003}. While the non-contextual relevance is constant, the contextual relevance changes every time $\mathcal{S}$ is modified. In theory, the only way to guarantee to select the optimal subset $\mathcal{S}^*$ is through exhaustive search wrappers. Exhaustive search is a very complex and time-consuming process because it expands exponentially with respect to the number of features \cite{ebiele_impact_2024, theng_feature_2024}; meaning there are exactly $2^{|\mathcal{F}|}-1$ possible combinations of features\footnote{$-1$ because the empty set $\emptyset$ has to be removed from the power set $\mathbb{P}(\mathcal{F})$ of $\mathcal{F}$. The power set $\mathbb{P}(\mathcal{F})$ of the set $\mathcal{F}$ is the set of all possible subsets of $\mathcal{F}$, including the empty set $\emptyset$ and $\mathcal{F}$ itself.}. Exhaustive search is not practical for high-dimensional datasets \cite{guyon_introduction_2003, ebiele_impact_2024}. 

As to FFS, it is a exploratory task meaning one needs to test multiple relevance (and redundancy in the case of multivariate selection) estimators. It also means testing multiple values of $k$. This exploratory approach gives a higher chance of getting a feature subset $\mathcal{S}$ that achieves a high predictive performance, but with higher computational cost. For instance, in their famous paper, \citet{guyon_introduction_2003} propose a check list on how to perform such exploration. 

In this paper, \textbf{we propose KGroups, a new versatile univariate mRMR algorithm that can easily be applied to any type of relevance estimator} i.e. supervised (such as mutual information or F-value), unsupervised (such as entropy or variance), embedded (such as random forest Gini importance). KGroups is fast and and parameterisable. KGroups is as fast as KBest and significantly faster than its multivariate counterpart of mRMR.

\section{Proposed Method}\label{sec:Proposed_Method}

This section provides a detailed description of the proposed univariate FFS algorithm based on clustering, KGroups. A binning clustering approach is chosen over classical ML clustering techniques (which will be explored in future studies) because it is an effective value-based univariate grouping method, computationally efficient, and parameterisable (see section \ref{sec:Experimental_Results}). Equation (\ref{eq:KGroups_general_formula}) is the general formulation of the KGroups algorithm. It consists of maximising the $\text{Rel}(F_i)$ within each feature cluster created using binning. 

\begin{align}
&\max_{b_j \in f(j;\alpha)} \text{Rel}(\mathcal{F}), \text{ subject to } \text{Rel}(\mathcal{F}) \leq b_j \label{eq:KGroups_general_formula},
\end{align}
where
\begin{align}
&f(j;\alpha)=\min \text{Rel}(\mathcal{F}) + ( \max \text{Rel}(\mathcal{F})-\min \text{Rel}(\mathcal{F}) )\times u_{j}^{\alpha}
\label{eq:KGroups_binning_formula}
\end{align}
and $u_{j} = \frac{j}{k}$, $j=1,2,..,k$; and $\alpha > 0$,  called power, is a real number used as the bin size regulation hyperparameter  (preferably $\alpha \in [0.05;3]$).

KGroups has three main steps: i) relevance estimation, ii) feature clustering based on their relevance values, and iii) actual feature selection. Each step is described below.

\begin{enumerate}
    \item \textbf{Relevance Estimation.} \label{item:Relevance EstimatiBoxplots of the executon}
    Like Max-Rel and mRMR, one has to select a relevance estimator and calculate the relevance $\text{Rel}(F_i)$ of each feature in the dataset. Any relevance estimator from the state of the art can be used (see section \ref{sec:Related_Work}). 

    \item \textbf{Feature Clustering.} \label{item:Feature Clustering}
    This step is built upon the last one and uses the relevance values to cluster the features into $k$ groups. Here, each feature $F_i$ is assigned to the cluster $C_j$ if and only if  $\text{Rel}(F_i) \in ]f(j-1;\alpha); f(j;\alpha)]$; noting that $f(0;\alpha) = \min \text{Rel}(\mathcal{F})$ (see Eq. (\ref{eq:KGroups_general_formula})). 
    
    Once the feature clustering is complete, we proceed to the actual feature selection step.

    \item \textbf{Actual Feature Selection.} \label{item:Actual Feature Selection}
    The goal here is to select the feature with the maximum relevance value $\text{Rel}(F_i)$ within each cluster $C_j$ for $j=1,2,..,k$. In practice, there is often more that one feature with the highest relevance value within a given cluster. There are many ways to address this issue. One is to return the first or last feature within the cluster with the highest  relevance value. Another way is to randomly select one or to just select all of the them. The first three approaches (select first, last or one randomly) have one major drawback which is reproducibility. If the features are shuffled then the resulting feature will be totally different. The second issue is explainability; why is that feature preferred over the others. As to selecting all the features with the highest relevance value, it fails to effective reduce the number of features when the the distribution of the relevance values is skewed. 
    
    To address all of the issues highlighted in the above paragraph, \textbf{we introduced the concept of tie-breaking. It is a sequential process which consists of using additional relevance estimators, one at a time, until either all the ties are broken or the list of tie-breakers is exhausted.} If all the ties are broken before the exhaustion of the list of tie-breakers, the process stops. On the other hand, if the list is exhausted and some ties are not broken, the remaining features are returned. This concept of tie-breaking adds a layer of robustness to the KGroups algorithm; at the cost of a little extra computation time.
\end{enumerate}

\section{Experimental Setup}\label{sec:Experimental_Setting}
The experiments are designed to evaluate the three feature selection algorithms i.e. KGroups (ours), mRMR, and KBest, ability to extract low-dimensional data from high-dimensional datasets, the robustness across classifiers of the low-dimensional data extraction, and the time complexity of the respective algorithms.

Our experiments employed 14 high-dimensional biological feature selection benchmark datasets (see Table \ref{tab:benchmark_datasets}), 6 widely used classifiers (with 4 highly reported in the FS literature i.e. KNeighbors, RandomForest, GaussianNB, and LinearSVC; see section \ref{subsec:classifiers} below), and 3 relevance estimators (mutual information, F-value, and Gini importance).

The experimental configuration includes KBest (Max-Rel), the two variants of mRMR (shown in Eqs. (\ref{eq:mRMR_subtraction_formula}) and (\ref{eq:mRMR_quotient_formula})), and seven variants of KGroups (for $\alpha\in \{0.3;0.5;0.7;1.0;1.3;1.5;1.7\}$; see Eqs. (\ref{eq:KGroups_general_formula}) and Eq. (\ref{eq:KGroups_binning_formula})) for mutual information and F-value relevance estimators. As to Gini importance relevance estimator, only one variant (the quotient variant proposed by \citet{zhao_maximum_2019}; see Eq. (\ref{eq:mRMR_quotient_formula})) of mRMR is available in public Python packages\footnote{\url{https://feature-engine.trainindata.com/en/latest/user_guide/selection/MRMR.html}} \cite{galli_feature-engine_2021}, thus only this one has been included. 

We performed hyperparameter finetuning for KGroups and mRMR algorithms. For KGroups, we finetuned the power hyperparameter $\alpha\in \{0.3;0.5;0.7;1.0;1.3;1.5;1.7\}$. For mRMR, we finetuned the operator between $\text{Rel}(F_i)$ and $\text{Red}(F_i, \mathcal{S})$; meaning between the difference (mRMR\_D, see Eq. (\ref{eq:mRMR_subtraction_formula})) and quotient (mRMR\_Q, see Eq. (\ref{eq:mRMR_quotient_formula})) formulations of mRMR. KGroups is more suitable for hyperparameter finetuning because it is fast compared to mRMR; in fact, on average (across the three relevance estimators) one KGroups variant is $122.8$ times faster than one of mRMR (see section \ref{subsec:Execution Time} for more details, especially Fig. \ref{fig:total_cpu_time} and Table \ref{tab:per_dataset_cpu_time}). In other words, \textbf{the time complexity of 123 variants of KGroups (123 different values of the the power hyperparameter $\alpha$) is equivalent to one of mRMR (mRMR\_D or mRMR\_Q)}. KBest has only one variant (no hyperparameters). 

\subsection{Classifiers} \label{subsec:classifiers}
The 6 classifiers used are $k$-Nearest Neighbors Classifier (KNeighbors), Gaussian Naive Bayes Classifier (GaussianNB), Random Forest Classifier (RandomForest), eXtreme Gradient Boosting Classifier (XGB), Multi-Layer Perceptron Classifier (MLP), and Linear Support Vector machine Classifier (LinearSVC). The classifiers are available in popular machine learning packages (XGB in XGBoost \cite{chen_xgboost_2016} package and the rest in the scikit-learn \cite{pedregosa_scikit-learn_2011} package). The classifiers are used with their default parameters and no hyper-parameter finetuning was performed. 
The short name of the classifiers between parenthesis has been automatically extracted from their respective Python packages. These short names are used when referencing the classifiers in the rest of this paper.

KNeighbors, RandomForest, GaussianNB, and LinearSVC have been chosen because they have been used in recent feature selection papers \cite{dubey_cosine-similarity_2017, yuan_cscim_fs_2023, yuan_feature_2025} and have been highly reported in the last two decades in FS literature. Another reason for selecting GaussianNB is because it is a probabilistic model. These are all widely used for classification tasks and are considered effective. 
XGB has been chosen because it is a robust tree-based model (compared to DecisionTree also known as classification and regression trees or CART) and MLP because it is a [simple] neural network (Feed Forward Network or FFS). These reasons aligned with the justifications given by \citet{ebiele_impact_2024} for selecting classifiers with different architectures and underlying learning principles. Choosing these classifiers will allow us to study the robustness of the FS methods and algorithms. It also simplifies the comparison with other FS methods.

\subsection{Datasets} \label{subsec:datasets}

The 14 high-dimensional biological benchmark datasets employed in the experiments are from \citet{li_feature_2017}'s feature selection repository. \citet{li_feature_2017}'s paper is a highly cited systematic review of feature selection literature which identified a total of 29 feature selection benchmark datasets including biological, text, face image, and hand-written image datasets. The 14 datasets selected for our experiments are all the biological benchmark datasets available on \citet{li_feature_2017}'s feature selection repository. In terms of tasks, there are binary and multiclass classification datasets with the maximum number of classes being 11 by the dataset Carcinom. The datasets with the lowest and highest dimensions are lung\_discrete and GLI\_85, respectively. In terms of sample size, these are all small datasets with the number of entries ranging from 50 to 203. SMK\_CAN\_187 is the largest dataset in terms of number of entries and features (dimensionality) combined with 187 entries and 19993 features. All these details about the datasets are summarised in Table \ref{tab:benchmark_datasets} below. The datasets are sorted alphabetically by name (case sensitive sorting).

This is very typical with tabular biological datasets, they are usually small size and high-dimensional \cite{bommert_benchmark_2022,li_feature_2017,yuan_feature_2025}. This type of datasets (small size and high-dimensional) are not eligible for deep neural networks (DNNs) because of their size. Additionally, \citet{grinsztajn_why_2022} have demonstrated that tree-based model (e.g. RandomForest, XGB) outperform DNNs (including Transformer models) on tabular data, with much less computational cost, due to irregular patterns in the target function, uninformative features, and non rotationally-invariant data where linear combinations of features misrepresent the information. The failure of DNNs on tabular data has been noticed before \citet{grinsztajn_why_2022} and gave birth to an entirely new research area called tabular deep learning. This new research area aims to stimulate efforts to design and implement DNNs architecture specialised on learning from tabular data.

\begin{table}[!htp]\centering
\caption{The list of benchmark datasets used in our experiments.}\label{tab:benchmark_datasets}
\begin{tabular}{lrrr}\toprule
\textbf{Name} &\textbf{\#Instances} &\textbf{\#Features} &\textbf{\#Classes} \\\midrule
ALLAML &72 &7129 &2 \\
CLL\_SUB\_111 &111 &11340 &3 \\
Carcinom &174 &9182 &11 \\
GLI\_85 &85 &22283 &2 \\
GLIOMA &50 &4434 &4 \\
Prostate\_GE &102 &5966 &2 \\
SMK\_CAN\_187 &187 &19993 &2 \\
TOX\_171 &171 &5748 &4 \\
colon &62 &2000 &2 \\
leukemia &72 &7070 &2 \\
lung &203 &3312 &5 \\
lung\_discrete &73 &325 &7 \\
lymphoma &96 &4026 &9 \\
nci9 &60 &9712 &9 \\
\bottomrule
\end{tabular}
\end{table}

\subsection{Training Setup and Evaluation} \label{subsec:Training Setup and Evaluation}

For $k \in [2,100]$, each classifier is trained using 5-fold cross validation. 

The classifiers have been evaluated using the accuracy metric for ease of comparison to related work. It is defined as:
\begin{align}
& \text{Accuracy} = \frac{\text{TP+TN}}{\text{TP+TN+FP+FN}} \label{eq:accuracy_formula}
\text{ (see confusion matrix in Table \ref{tab:confusion_matrix})}
\end{align}

The classifiers were trained on a shared high performance computing (HPC) node running a Red Hat Enterprise Linux (RHEL) 9.5 machine with 356 GB RAM and an Intel Xeon Gold 6152 CPU (30.25 MB L3 Cache and 2.1 up to 3.7 GHz Frequency). 

The datasets have been scaled using the scikit-learn Standard scaler before training. For KGroups, \textbf{cosine similarity has been employed as tie-breaker for Mutual information and Mutual information has been employed as tie-breaker for F-value and Gini importance.} 

\begin{table}[!h]
\caption{Confusion matrix.}
\label{tab:confusion_matrix}
\centering
\begin{tabular}{cc|c|c|c}
\hhline{~~|--}
& & \multicolumn{2}{ |c| }{Actual} \\ 
\hhline{~~|--}
& & Positive (P) & Negative (N) \\ 
\hhline{----}
\multicolumn{1}{ |c  }{\multirow{2}{*}{Predicted} } &
\multicolumn{1}{ |c| }{Positive (P)} & True Positive (TP) & False Positive (FP) &  \\ \hhline{~|---}
\multicolumn{1}{ |c  }{}                        &
\multicolumn{1}{ |c| }{Negative (N)} & False Negative (FN) & True Negative (TN) &  \\ \hhline{----}
\end{tabular}
\end{table}

\section{Experimental Results and Discussion}\label{sec:Experimental_Results}
The experimental results are shown and discussed below for a set of 3 FS related tasks. KGroups, KBest, and mRMR algorithms have been compared on low-dimension feature extraction, their robustness across heterogeneous ML models, and execution time. To make comparison easier to understand and follow, we performed a pairwise comparision of the three FFS algorithms; meaning KGroups versus KBest, then KGroups verus mRMR for the low-dimension feature extraction task and the robustness across heterogeneous ML models of the extraction. 

\subsection{Low-dimension Feature Extraction}
The low-dimension feature extraction task consists of selecting a subset of $k$ features from the original feature set such that $k\ll|\mathcal{F}|$; here we chose $k=100$. Therefore, for $k \in [2,100]$, we trained 6 different classifiers using 5-fold cross validation and recorded the average and standard deviation accuracy score. 

Table \ref{tab:Best overall configs - KGvsKB} displays the best overall experimental configuration per relevance estimator and dataset as a result of systematic search. The results reported below are the best configurations among 594 possible results per variant per dataset (594 = 99 possible value of $k \text{ }\times$ 6 classifiers). For mRMR and KGroups, the variant (after hyperparameter finetuning) with the highest accuracy per dataset is reported.

\textbf{KGroups versus KBest:} KGroups with Gini importance relevance estimator got the best configuration on 7 datasets.  KGroups with mutual information estimator got the best configuration on 9 datasets. KGroups with F-value relevance estimator got the best configuration on 10 datasets. KGroups and KBest tied on 4, 3, and 1 datasets, for Gini importance, mutual information, and F-value relevance estimator, respectively. Across the three relevance estimators, KGroups achieved best configuration on 11 datasets overall, tying with KBest on 4 datasets (7 excluding ties). KBest achieved best configuration on 2 datasets excluding ties. Both, KGroups and KBest failed to achieved the best configuration on the TOX\_171 dataset.  

\textbf{KGroups versus mRMR:} KGroups with Gini importance relevance estimator got the best configuration on 9 datasets.  KGroups with mutual information estimator got the best configuration on 5 datasets. KGroups with F-value relevance estimator got the best configuration on 2 datasets. KGroups and mRMR tied on 2, 2, and 4 datasets, for Gini importance, mutual information, and F-value relevance estimator, respectively. Across the three relevance estimators, KGroups achieved best configuration on 6 datasets overall, tying with mRMR on 2 datasets (4 excluding ties). mRMR achieved best configuration on 7 datasets excluding ties. Both, KGroups and mRMR failed to achieved the best configuration on the TOX\_171 dataset.  

Figures \ref{fig:num_selected_for_gini}-\ref{fig:num_selected_for_Fvalue} below show the distribution of the number of features selected by the best configurations per datasets and per FS algorithms (and their respective variants). Overall, KGroups (and its variants) select fewer features than the two mRMR variants and KBest. This is particularly visible for Gini importance and F-value relevance estimators. As to the mutual information relevance estimator, some of KGroups variants occasionally select more features than at least one mRMR variants or KBest. This is noticeable with the colon dataset on Fig. \ref{fig:num_selected_for_mutual_info}. However, there are no instances where the two mRMR variants and KBest select fewer features than all the variants of KGroups.

It is worth mentioning that the distribution of the number of selected features by KGroups variants for F-value relevance estimator on the nci9 dataset is abnormal and will be discussed in section \ref{sec:KGroups Limitations and Potentials} below. 

\begin{table}[!htp]\centering
\caption{Best overall configurations of \textbf{KBest and KGroups} across classifiers. The highest accuracy per relevance estimator excluding ties (Win) is in \textbf{bold}. Tied highest accuracy scores (Draw) are \underline{underlined}. The star ($^{*}$) indicates best configuration per dataset. $^{\text{a}}$ indicates abnormally low accuracy.
}
\label{tab:Best overall configs - KGvsKB}
\begin{tabular}{l|r|rr|rr|rrr}\toprule
& &\multicolumn{2}{c}{Gini importance} &\multicolumn{2}{|c|}{Mutual info.} &\multicolumn{2}{c}{F-value} \\\cmidrule{3-8}
\textbf{Dataset} &\textbf{All feat.} &\textbf{KBest} &\textbf{KGroups} &\textbf{KBest} &\textbf{KGroups} &\textbf{KBest} &\textbf{KGroups} \\\midrule
ALLAML &95.91 &\ul{100.00*} &\ul{100.00*} &98.57 &\textbf{100.00*} &98.57 &\textbf{100.00*} \\
CLL\_SUB\_111 &73.82 &\textbf{88.26*} &88.22 &\ul{84.66} &\ul{84.66} &71.15 &\textbf{79.21} \\
Carcinom &95.47 &94.82 &\textbf{96.54*} &95.38 &\textbf{95.39} &\textbf{92.52} &77.01 \\
GLIOMA &78.00 &\textbf{98.00*} &96.00 &90.00 &\textbf{94.00} &88.00 &\textbf{90.00} \\
GLI\_85 &84.62 &95.29 &\textbf{98.82*} &96.47 &\textbf{98.82*} &95.29 &\textbf{97.65} \\
Prostate\_GE &89.68 &\textbf{97.00} &96.10 &96.05 &\textbf{97.05*} &95.10 &\textbf{97.05*} \\
SMK\_CAN\_187 &70.18 &76.49 &\textbf{78.61*} &\textbf{76.49} &75.43 &\textbf{75.39} &74.82 \\
TOX\_171 &91.15* &83.09 &\textbf{84.29} &69.03 &\textbf{73.66} &70.20 &\textbf{76.00} \\
colon &84.74 &90.00 &\textbf{91.79} &93.33 &\textbf{95.00*} &90.13 &\textbf{93.59} \\
leukemia &97.73 &\ul{100.00*} &\ul{100.00*} &\ul{100.00*} &\ul{100.00*} &\ul{100.00*} &\ul{100.00*} \\
lung &94.10 &97.04 &\textbf{98.01*} &97.52 &\textbf{97.54} &94.60 &\textbf{96.55} \\
lung\_discrete &89.55 &93.14 &\textbf{95.90*} &\textbf{91.71} &90.48 &93.14 &\textbf{94.38} \\
lymphoma &91.38 &\ul{95.84*} &\ul{95.84*} &\ul{95.79} &\ul{94.79} &87.53 &\textbf{94.79} \\
nci9 &52.22$^{\text{a}}$ &\ul{78.33*} &\ul{78.33*} &48.33$^{\text{a}}$ &\textbf{66.67} &\textbf{78.33*} &23.33$^{\text{a}}$ \\
\hline\hline
Win / Draw & &3 / 4 &7 / 4 &2 / 3 &9 / 3 &3 / 1 &10 / 1 \\
\bottomrule
\end{tabular}
\end{table}

\begin{table}[!htp]\centering
\caption{Best overall configurations of \textbf{mRMR and KGroups} across classifiers. The highest accuracy per relevance estimator excluding ties (Win) is in \textbf{bold}. Tied highest accuracy scores (Draw) are \underline{underlined}. The star ($^{*}$) indicates best configuration per dataset. $^{\text{a}}$ indicates abnormally low accuracy.
}
\label{tab:Best overall configs - KGvsMRMR}
\begin{tabular}{l|r|rr|rr|rrr}\toprule
& &\multicolumn{2}{c}{Gini importance} &\multicolumn{2}{|c|}{Mutual info.} &\multicolumn{2}{c}{F-value} \\\cmidrule{3-8}
\textbf{Dataset} &\textbf{All feat.} &\textbf{mRMR} &\textbf{KGroups} &\textbf{mRMR} &\textbf{KGroups} &\textbf{mRMR} &\textbf{KGroups} \\\midrule
ALLAML &95.91 &\ul{100.00*} &\ul{100.00*} &\ul{100.00*} &\ul{100.00*} &\ul{100.00*} &\ul{100.00*} \\
CLL\_SUB\_111 &73.82 &\textbf{91.86*} &88.22 &\textbf{90.04} &84.66 &\textbf{85.57} &79.21 \\
Carcinom &95.47 &95.98 &\textbf{96.54} &\textbf{96.55*} &95.39 &\textbf{95.97} &77.01 \\
GLIOMA &78.00 &94.00 &\textbf{96.00*} &90.00 &\textbf{94.00} &88.00 &\textbf{90.00} \\
GLI\_85 &84.62 &96.47 &\textbf{98.82*} &97.65 &\textbf{98.82*} &\ul{97.65} &\ul{97.65} \\
Prostate\_GE &89.68 &96.05 &\textbf{96.10} &\textbf{98.00*} &97.05 &\ul{97.05} &\ul{97.05} \\
SMK\_CAN\_187 &70.18 &76.49 &\textbf{78.61} &73.81 &\textbf{75.43} &\textbf{81.28*} &74.82 \\
TOX\_171 &91.15* &\textbf{85.95} &84.29 &\textbf{77.78} &73.66 &\textbf{80.13} &76.00 \\
colon &84.74 &90.00 &\textbf{91.79} &93.46 &\textbf{95.00*} &93.33 &\textbf{93.59} \\
leukemia &97.73 &\ul{100.00*} &\ul{100.00*} &\ul{100.00*} &\ul{100.00*} &\ul{100.00*} &\ul{100.00*} \\
lung &94.10 &97.04 &\textbf{98.01} &\textbf{98.02*} &97.54 &\textbf{97.52} &96.55 \\
lung\_discrete &89.55 &94.57 &\textbf{95.90*} &\textbf{91.81} &90.48 &\textbf{94.57} &94.38 \\
lymphoma &91.38 &\textbf{98.95*} &95.84 &\textbf{96.84} &94.79 &\textbf{97.89} &94.79 \\
nci9 &52.22$^{\text{a}}$ &76.67 &\textbf{78.33} &46.67$^{\text{a}}$ &\textbf{66.67} &\textbf{85.00*} &23.33$^{\text{a}}$ \\
\hline\hline
Win / Draw & &3 / 2 &9 / 2 &7 / 2 &5 / 2 &8 / 4 &2 / 4 \\
\bottomrule
\end{tabular}
\end{table}

\begin{figure}[!htp]
\centering
\includegraphics[width=0.99\textwidth, height=8cm]{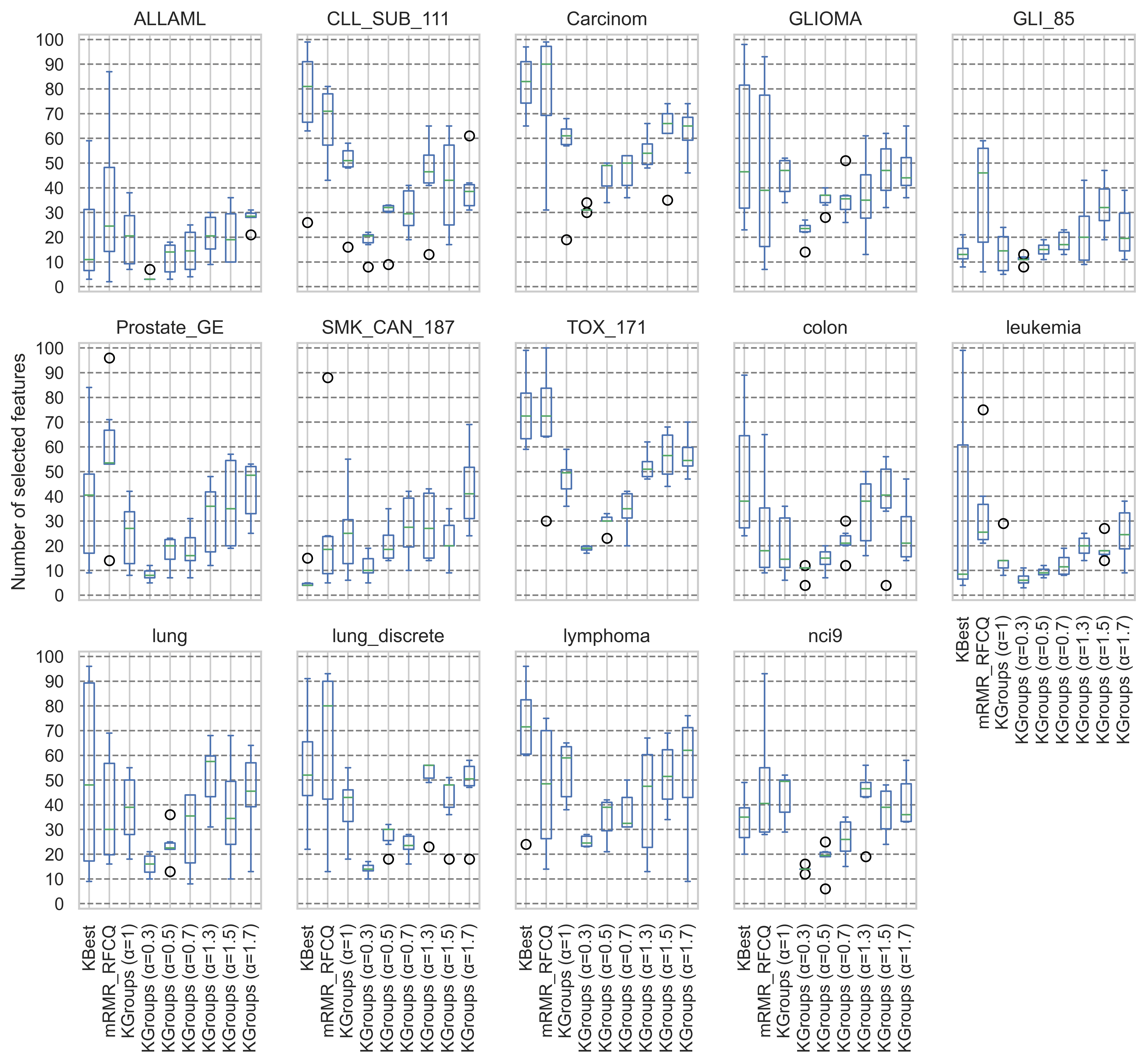}
\caption{Distribution of number of features selected by KBest, mRMR, and KGroups using \textbf{Gini importance} relevance estimator and KNeighbors, GaussianNB, RandomForest, XGB, MLP, and LinearSVC classification models.}\label{fig:num_selected_for_gini}
\end{figure}

\begin{figure}[!htp]
\centering
\includegraphics[width=0.99\textwidth, height=8cm]{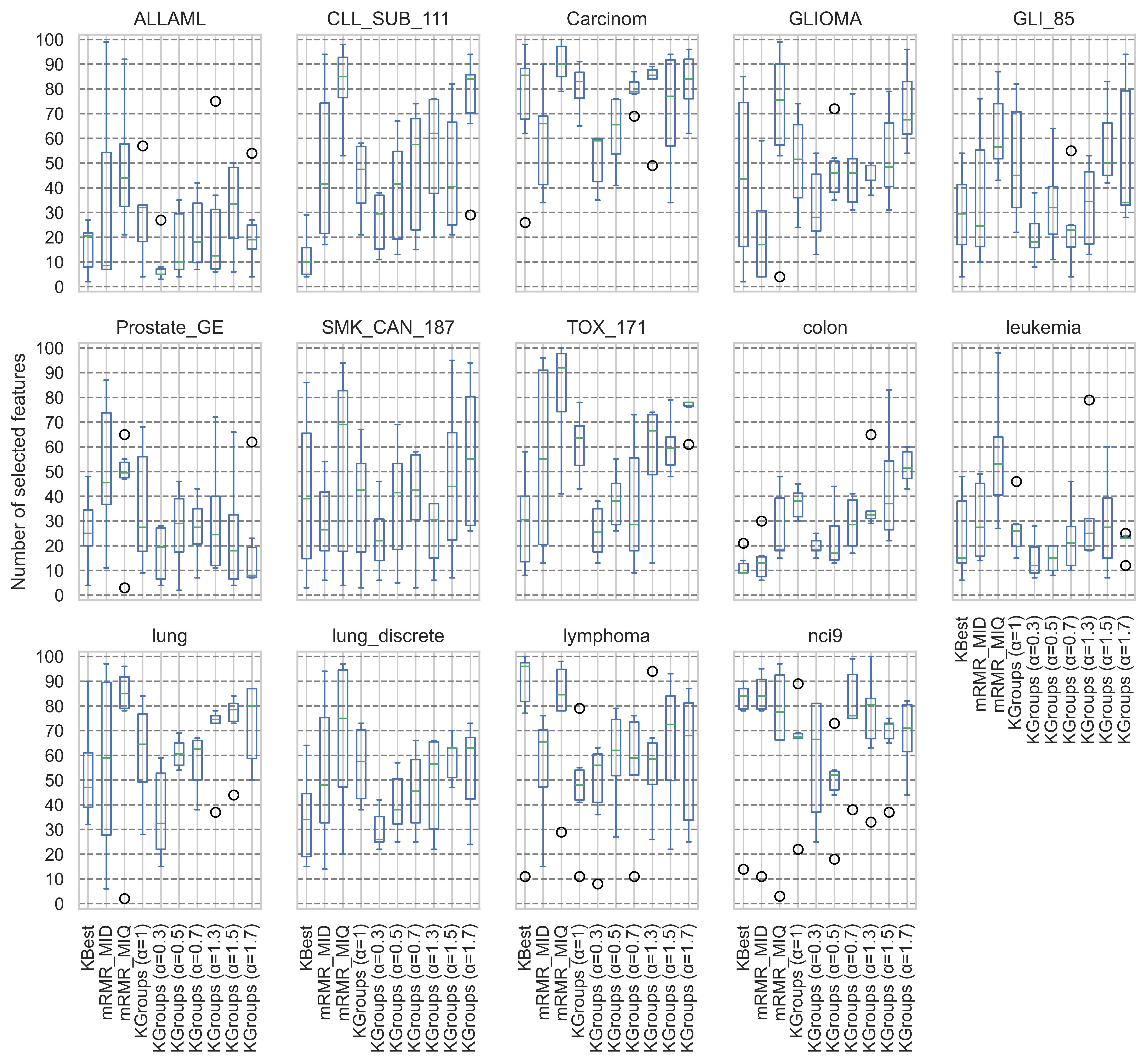}
\caption{Distribution of number of features selected by KBest, mRMR, and KGroups using \textbf{mutual information} relevance estimator and KNeighbors, GaussianNB, RandomForest, XGB, MLP, and LinearSVC classification models.}\label{fig:num_selected_for_mutual_info}
\end{figure}

\begin{figure}[!htp]
\centering
\includegraphics[width=0.99\textwidth, height=8cm]{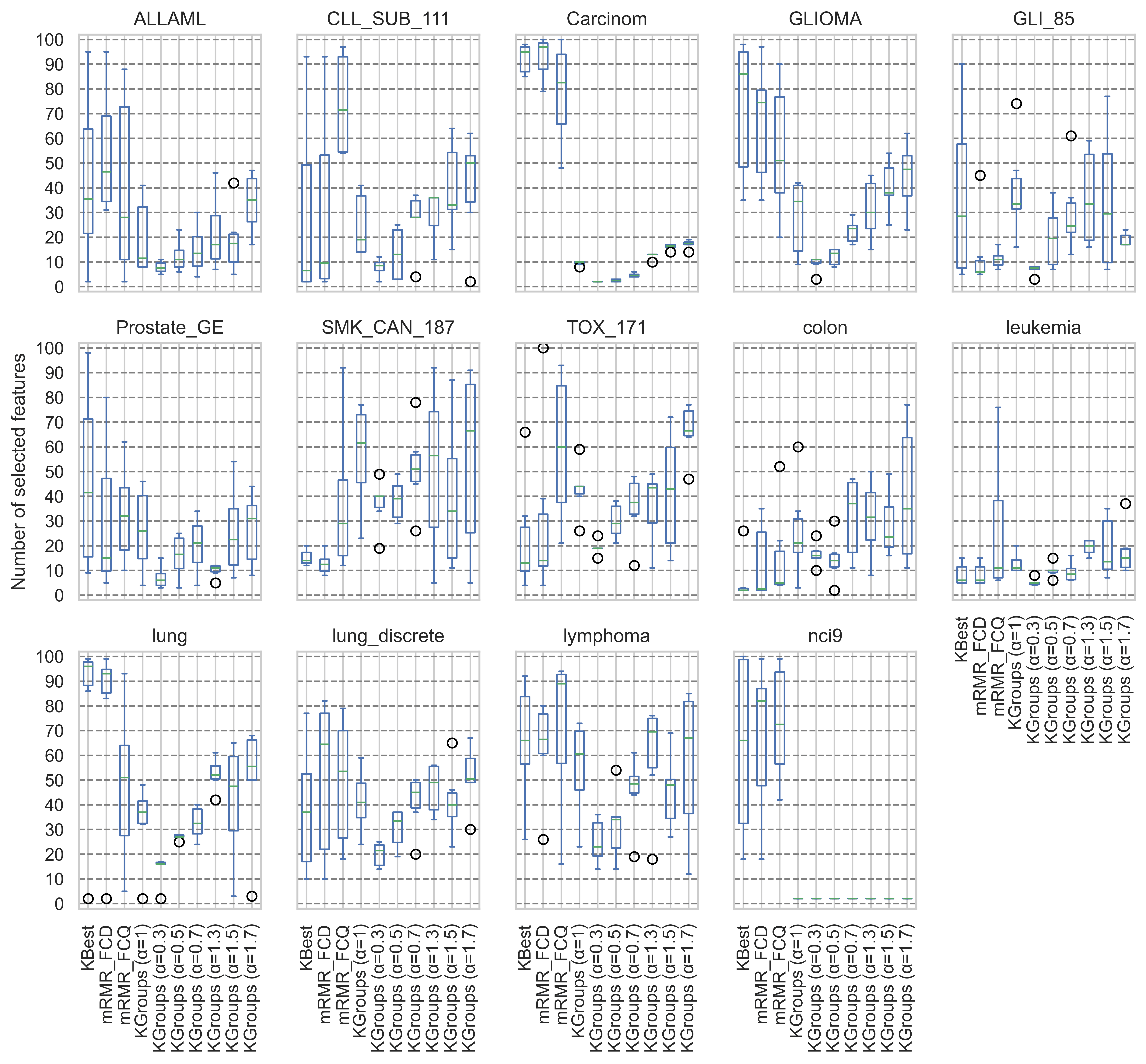}
\caption{Distribution of number of features selected by KBest, mRMR, and KGroups using \textbf{F-value} relevance estimator and KNeighbors, GaussianNB, RandomForest, XGB, MLP, and LinearSVC classification models.}\label{fig:num_selected_for_Fvalue}
\end{figure}

\subsection{Robustness of Feature Extraction}
The FS task involves many choices that can each impact the resulting feature subset considerably as discussed in \citet{guyon_introduction_2003}. One of those choices is the classifier used; another is the number features $k$ to be returned (this can be set manually or systematically given a range of possible values of $k$ as we did). We examined 6 popular classifiers, each trained independently and recorded the best configurations for KBest, mRMR and KGRoups using the three relevance estimators (Gini importance, mutual information and F-value). Calculating the average (and standard deviation) accuracy of the top configurations per classifier allows us to study the robustness of the feature selection algorithms. High average accuracy means the algorithm is reliable in returning feature subset with the highest accuracy independently of the classifier. Low standard deviation indicates high convergence across classifiers; another sign of reliability. Tables \ref{tab:Average best configs - KGvsKB} and \ref{tab:Average best configs - KGvsMRMR} highlight the results of the studies of the robustness of the three feature selection algorithms included in this paper. 

\textbf{KGroups versus KBest:} KGroups with Gini importance relevance estimator got the best configuration on 12 datasets. KGroups with mutual information estimator got the best configuration on 13 datasets. KGroups with F-value relevance estimator got the best configuration on 11 datasets. KGroups and KBest have not tied on any of the three relevance estimators. Across the three relevance estimators, KGroups achieved best configuration on 13 datasets (excluding ties). 
Both, KGroups and KBest failed to achieved the best configuration on the TOX\_171 dataset.  

\textbf{KGroups versus mRMR:} KGroups with Gini importance relevance estimator got the best configuration on 8 datasets.  KGroups with mutual information estimator got the best configuration on 7 datasets. KGroups with F-value relevance estimator got the best configuration on 5 datasets. KGroups and mRMR tied on 2, 0, and 0 datasets, for Gini importance, mutual information, and F-value relevance estimator, respectively. Across the three relevance estimators, KGroups achieved best configuration on 7 datasets overall, tying with mRMR on 2 datasets (5 excluding ties). mRMR achieved best configuration on 7 datasets excluding ties. 

Another thing noticeable when looking at the boxplots on Figs. \ref{fig:num_selected_for_gini}-\ref{fig:num_selected_for_Fvalue} above is their difference in range ($\max - \min$, excluding outliers). In general, KGroups variants boxplots are narrower than KBest and the two variants of mRMR across the three relevance estimators, especially for Gini importance and F-value. In other words, the numbers of keys features selected by KGroups vary little from one classifier to another. This is a additional sign of robustness of the KGroups algorithm.

\begin{table}[!htp]
\centering
\small
\setlength{\tabcolsep}{3.5pt}
\caption{Average $\pm$ standard deviation of the best configurations of \textbf{KBest and KGroups (KG)} across classifiers. The highest average accuracy per relevance estimator excluding ties (Win) is in \textbf{bold}. Tied highest average accuracy scores (Draw) are \underline{underlined}. The star ($^{*}$) indicates best configuration per dataset. $^{\text{a}}$ indicates abnormally low accuracy.}
\label{tab:Average best configs - KGvsKB}
\begin{tabular}{@{}l|
>{\raggedleft\arraybackslash}p{0.95cm}|
*{2}{>{\raggedleft\arraybackslash}p{0.95cm}}|
*{2}{>{\raggedleft\arraybackslash}p{0.95cm}}|
*{2}{>{\raggedleft\arraybackslash}p{0.95cm}}@{}}
\toprule
& & \multicolumn{2}{c}{Gini import.} & \multicolumn{2}{|c|}{Mutual info.} & \multicolumn{2}{c}{F-value} \\
\cmidrule(lr){3-4}\cmidrule(lr){5-6}\cmidrule(lr){7-8}
\textbf{Dataset} & \textbf{All feat.} & \textbf{KBest} & \textbf{KG} & \textbf{KBest} & \textbf{KG} & \textbf{KBest} & \textbf{KG} \\
\midrule
ALLAML & 87.88 & 98.36 & \textbf{99.56*} & 97.43 & \textbf{99.33} & 96.98 & \textbf{98.33} \\
& $\pm$7.31 & $\pm$1.37 & $\pm$1.09 & $\pm$2.21 & $\pm$1.12 & $\pm$2.03 & $\pm$1.41 \\
\addlinespace
CLL\_SUB\_111 & 65.39 & 85.28 & \textbf{85.99*} & 78.96 & \textbf{80.00} & 65.62 & \textbf{73.36} \\
& $\pm$6.38 & $\pm$2.50 & $\pm$1.68 & $\pm$4.41 & $\pm$2.64 & $\pm$4.58 & $\pm$4.19 \\
\addlinespace
Carcinom & 85.13 & 86.88 & \textbf{90.41*} & 87.26 & \textbf{89.82} & \textbf{81.93} & 69.05 \\
& $\pm$7.27 & $\pm$9.89 & $\pm$6.06 & $\pm$10.71 & $\pm$6.61 & $\pm$14.66 & $\pm$10.44 \\
\addlinespace
GLIOMA & 62.67 & 90.33 & \textbf{90.67*} & 85.67 & \textbf{87.67} & 78.67 & \textbf{84.67} \\
& $\pm$16.54 & $\pm$6.74 & $\pm$2.42 & $\pm$3.44 & $\pm$4.97 & $\pm$6.41 & $\pm$3.72 \\
\addlinespace
GLI\_85 & 77.63 & 94.90 & \textbf{96.27} & 94.90 & \textbf{96.86*} & 93.72 & \textbf{95.10} \\
& $\pm$8.39 & $\pm$0.60 & $\pm$0.89 & $\pm$1.42 & $\pm$0.61 & $\pm$0.96 & $\pm$0.88 \\
\addlinespace
Prostate & 81.34 & \textbf{95.23} & 94.61 & 95.41 & \textbf{95.72*} & 94.60 & \textbf{95.42} \\
& $\pm$11.16 & $\pm$1.29 & $\pm$1.03 & $\pm$0.79 & $\pm$0.81 & $\pm$0.51 & $\pm$0.80 \\
\addlinespace
SMK\_CAN\_187 & 67.52 & 74.59 & \textbf{75.68*} & 71.66 & \textbf{73.43} & \textbf{74.83} & 72.80 \\
& $\pm$2.57 & $\pm$1.44 & $\pm$0.80 & $\pm$3.08 & $\pm$1.25 & $\pm$0.67 & $\pm$1.33 \\
\addlinespace
TOX\_171 & 78.94* & \textbf{77.50} & 76.83 & 66.47 & \textbf{69.41} & 66.49 & \textbf{71.21} \\
& $\pm$7.46 & $\pm$3.91 & $\pm$2.56 & $\pm$2.70 & $\pm$2.10 & $\pm$2.62 & $\pm$2.85 \\
\addlinespace
colon & 77.11 & 87.93 & \textbf{90.11} & 91.26 & \textbf{91.90*} & 89.10 & \textbf{91.48} \\
& $\pm$6.14 & $\pm$1.61 & $\pm$0.94 & $\pm$1.68 & $\pm$2.01 & $\pm$0.80 & $\pm$1.31 \\
\addlinespace
leukemia & 87.88 & 99.56 & \textbf{99.78*} & 99.08 & \textbf{99.78*} & 98.89 & \textbf{99.78*} \\
& $\pm$6.98 & $\pm$0.69 & $\pm$0.54 & $\pm$1.69 & $\pm$0.54 & $\pm$1.00 & $\pm$0.54 \\
\addlinespace
lung & 84.02 & 95.72 & \textbf{96.47*} & 94.98 & \textbf{95.97} & 92.03 & \textbf{93.75} \\
& $\pm$15.75 & $\pm$1.69 & $\pm$0.65 & $\pm$2.91 & $\pm$1.48 & $\pm$4.93 & $\pm$4.29 \\
\addlinespace
lung\_discrete & 77.58 & 88.35 & \textbf{89.02*} & \textbf{87.11} & 86.79 & 87.52 & \textbf{87.70} \\
& $\pm$15.61 & $\pm$7.97 & $\pm$5.92 & $\pm$6.74 & $\pm$4.39 & $\pm$6.02 & $\pm$7.31 \\
\addlinespace
lymphoma & 75.06 & 86.70 & \textbf{89.15} & 89.13 & \textbf{89.15} & 82.90 & \textbf{89.62*} \\
& $\pm$19.38 & $\pm$11.29 & $\pm$7.27 & $\pm$8.62 & $\pm$5.69 & $\pm$6.91 & $\pm$6.07 \\
\addlinespace
nci9 & 43.33$^{\text{a}}$ & 68.33 & \textbf{70.56} & 41.11$^{\text{a}}$ & \textbf{57.50} & \textbf{71.39*} & 22.50$^{\text{a}}$ \\
& $\pm$8.80 & $\pm$9.25 & $\pm$6.97 & $\pm$8.21 & $\pm$7.66 & $\pm$5.91 & $\pm$1.39 \\
\midrule
Win/Draw & & 2/0 & 12/0 & 1/0 & 13/0 & 3/0 & 11/0 \\
\bottomrule
\end{tabular}
\end{table}

\begin{table}[!htp]
\centering
\small
\setlength{\tabcolsep}{3.5pt}
\caption{Average $\pm$ standard deviation of the best configurations of \textbf{mRMR and KGroups (KG)} across classifiers. The highest average accuracy per relevance estimator excluding ties (Win) is in \textbf{bold}. Tied highest average accuracy scores (Draw) are \underline{underlined}. The star ($^{*}$) indicates best configuration per dataset. $^{\text{a}}$ indicates abnormally low accuracy.}
\label{tab:Average best configs - KGvsMRMR}
\begin{tabular}{@{}l|
>{\raggedleft\arraybackslash}p{0.95cm}|
*{2}{>{\raggedleft\arraybackslash}p{0.95cm}}|
*{2}{>{\raggedleft\arraybackslash}p{0.95cm}}|
*{2}{>{\raggedleft\arraybackslash}p{0.95cm}}@{}}
\toprule
& & \multicolumn{2}{c}{Gini import.} & \multicolumn{2}{|c|}{Mutual info.} & \multicolumn{2}{c}{F-value} \\
\cmidrule(lr){3-4}\cmidrule(lr){5-6}\cmidrule(lr){7-8}
\textbf{Dataset} & \textbf{All feat.} & \textbf{mRMR} & \textbf{KG} & \textbf{mRMR} & \textbf{KG} & \textbf{mRMR} & \textbf{KG} \\
\midrule
ALLAML & 87.88 & 98.16 & \textbf{99.56*} & 99.10 & \textbf{99.33} & 98.13 & \textbf{98.33} \\
& $\pm$7.31 & $\pm$1.45 & $\pm$1.09 & $\pm$0.70 & $\pm$1.12 & $\pm$2.11 & $\pm$1.41 \\
\addlinespace
CLL\_SUB\_111 & 65.39 & \textbf{88.12*} & 85.99 & \textbf{85.14} & 80.00 & \textbf{81.05} & 73.36 \\
& $\pm$6.38 & $\pm$2.75 & $\pm$1.68 & $\pm$3.35 & $\pm$2.64 & $\pm$5.67 & $\pm$4.19 \\
\addlinespace
Carcinom & 85.13 & \ul{90.41*} & \ul{90.41*} & \textbf{90.22} & 89.82 & \textbf{89.82} & 69.05 \\
& $\pm$7.27 & $\pm$5.96 & $\pm$6.06 & $\pm$8.84 & $\pm$6.61 & $\pm$10.71 & $\pm$10.44 \\
\addlinespace
GLIOMA & 62.67 & 88.00 & \textbf{90.67*} & 83.00 & \textbf{87.67} & 83.33 & \textbf{84.67} \\
& $\pm$16.54 & $\pm$5.66 & $\pm$2.42 & $\pm$7.77 & $\pm$4.97 & $\pm$4.32 & $\pm$3.72 \\
\addlinespace
GLI\_85 & 77.63 & 94.70 & \textbf{96.27} & 94.71 & \textbf{96.86*} & \textbf{95.69} & 95.10 \\
& $\pm$8.39 & $\pm$2.07 & $\pm$0.89 & $\pm$2.44 & $\pm$0.61 & $\pm$1.92 & $\pm$0.88 \\
\addlinespace
Prostate & 81.34 & \textbf{94.92} & 94.61 & \textbf{95.74*} & 95.72 & 95.25 & \textbf{95.42} \\
& $\pm$11.16 & $\pm$0.95 & $\pm$1.03 & $\pm$0.79 & $\pm$0.81 & $\pm$1.14 & $\pm$0.80 \\
\addlinespace
SMK\_CAN\_187 & 67.52 & 74.22 & \textbf{75.68} & 69.96 & \textbf{73.43} & \textbf{78.59*} & 72.80 \\
& $\pm$2.57 & $\pm$1.76 & $\pm$0.80 & $\pm$2.23 & $\pm$1.25 & $\pm$2.25 & $\pm$1.33 \\
\addlinespace
TOX\_171 & 78.94 & \textbf{80.81*} & 76.83 & \textbf{74.27} & 69.41 & \textbf{75.68} & 71.21 \\
& $\pm$7.46 & $\pm$4.00 & $\pm$2.56 & $\pm$2.77 & $\pm$2.10 & $\pm$2.38 & $\pm$2.85 \\
\addlinespace
colon & 77.11 & 88.22 & \textbf{90.11} & 91.86 & \textbf{91.90*} & 90.90 & \textbf{91.48} \\
& $\pm$6.14 & $\pm$1.78 & $\pm$0.94 & $\pm$2.01 & $\pm$2.01 & $\pm$1.99 & $\pm$1.31 \\
\addlinespace
leukemia & 87.88 & \ul{99.78*} & \ul{99.78*} & 99.10 & \textbf{99.78*} & 98.89 & \textbf{99.78*} \\
& $\pm$6.98 & $\pm$0.54 & $\pm$0.54 & $\pm$1.65 & $\pm$0.54 & $\pm$1.00 & $\pm$0.54 \\
\addlinespace
lung & 84.02 & 95.56 & \textbf{96.47*} & \textbf{96.06} & 95.97 & \textbf{95.80} & 93.75 \\
& $\pm$15.75 & $\pm$1.76 & $\pm$0.65 & $\pm$1.68 & $\pm$1.48 & $\pm$2.15 & $\pm$4.29 \\
\addlinespace
lung\_discrete & 77.58 & 88.98 & \textbf{89.02} & \textbf{87.89} & 86.79 & \textbf{89.70*} & 87.70 \\
& $\pm$15.61 & $\pm$6.46 & $\pm$5.92 & $\pm$5.16 & $\pm$4.39 & $\pm$6.31 & $\pm$7.31 \\
\addlinespace
lymphoma & 75.06 & \textbf{91.31*} & 89.15 & \textbf{90.69} & 89.15 & \textbf{90.18} & 89.62 \\
& $\pm$19.38 & $\pm$8.96 & $\pm$7.27 & $\pm$6.93 & $\pm$5.69 & $\pm$8.96 & $\pm$6.07 \\
\addlinespace
nci9 & 43.33$^{\text{a}}$ & 70.00 & \textbf{70.56} & 40.00$^{\text{a}}$ & \textbf{57.50} & \textbf{76.11*} & 22.50$^{\text{a}}$ \\
& $\pm$8.80 & $\pm$10.06 & $\pm$6.97 & $\pm$7.82 & $\pm$7.66 & $\pm$8.41 & $\pm$1.39 \\
\midrule
Win/Draw & & 4/2 & 8/2 & 7/0 & 7/0 & 9/0 & 5/0 \\
\bottomrule
\end{tabular}
\end{table}

\subsection{Execution Time}\label{subsec:Execution Time}
In this section, we study the time complexity of the three feature selection algorithms during the feature extraction using their respective Central Processing Unit (CPU) execution time. CPU time is the total duration any of the CPU cores spends actively executing instructions for a specific task or program. This excludes system interruptions, time spent waiting for input or output operations, network requests, or the system entering a low-power or idle state. CPU time is equivalent for sequential, parallel or distributed tasks or programs. For our experiments, we recorded both CPU and Wall times, and they are equivalent. However, the CPU time has been chosen over Wall time to be reported below for the reasons mentioned above. 

Figure \ref{fig:total_cpu_time} and Table \ref{tab:per_dataset_cpu_time} below display the CPU time in seconds of three FS algorithms across the three relevance estimators. Time complexity of mRMR variants are equivalent so are KGroups variants, for a given relevance and redundancy estimators. The time complexity of KBest and KGroups variants is equivalent for different relevance estimators e.g. Gini importance, mutual information, and F-value (see Fig. \ref{fig:total_cpu_time} below). However, mRMR time complexity heavily depends on the time complexity of the redundancy estimator. That is why the time complexity of mRMR is the same for Gini importance and F-value but dramatically different for mutual information. In fact, time complexity of mRMR is on average $8.2$ times slower for mutual information compared to  Gini importance and F-value. These results aligned with the one reported by \citet{zhao_maximum_2019} (see section \ref{subsec:Multivariate FFS}).

Table \ref{tab:per_dataset_cpu_time} shows the CPU execution time in seconds of the three algorithms per dataset and relevance estimator. Overall, KGroups is slightly faster that KBest and significantly faster than mRMR. One can notice that the execution time per dataset of KGroups and KBest changes little across the three relevance estimator; confirming that the choice of the relevance estimator has limited effects on the time complexity of these two algorithms. However, the choice of the relevance and particularly of the redundancy estimators has a high impact on the time complexity of mRMR algorithm. 

\begin{figure}[!htp]
\centering
\includegraphics[width=0.9\textwidth, height=6cm]{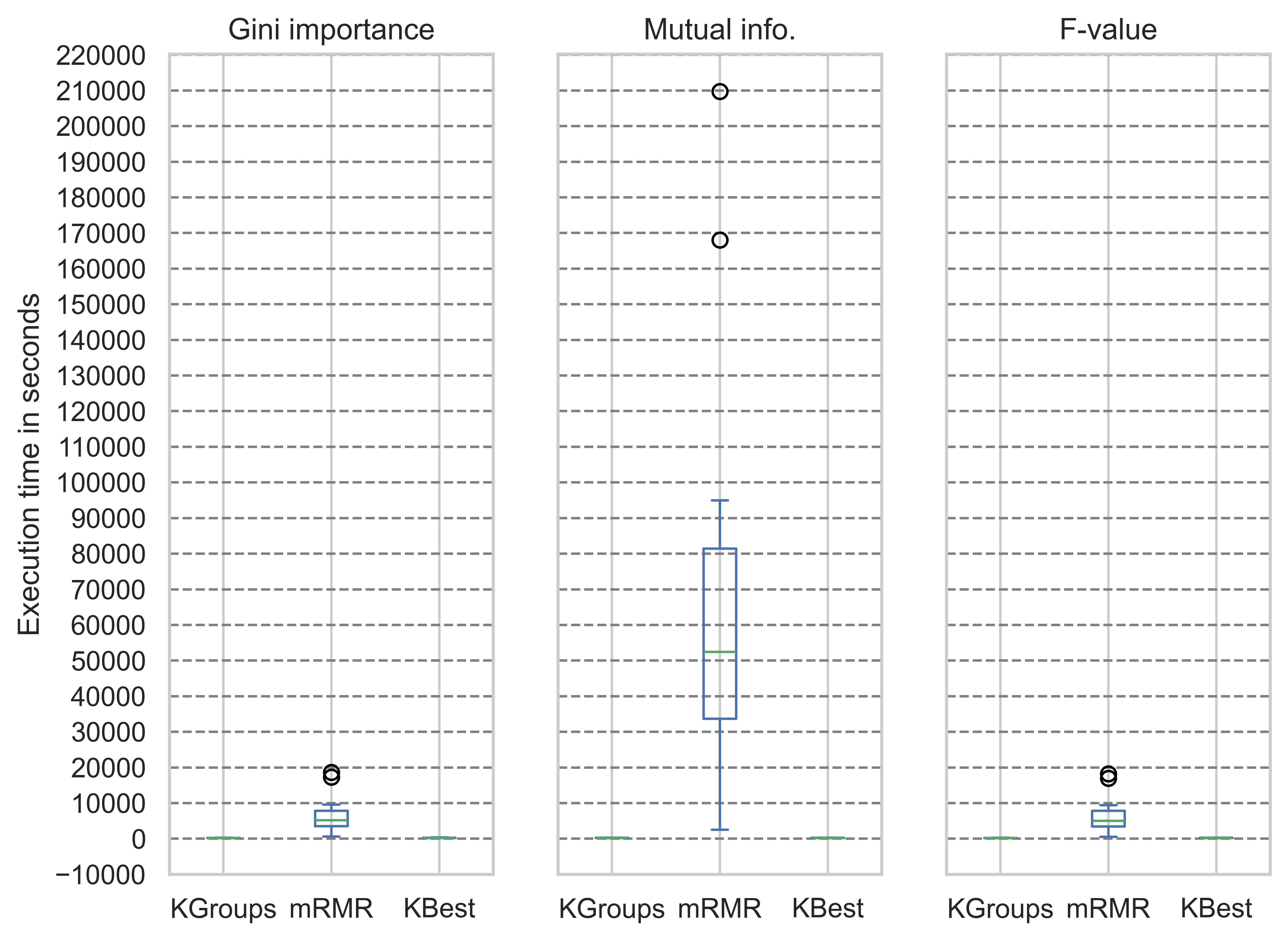}
\caption{Boxplots of the execution time in seconds of KBest, mRMR, and KGroups.}\label{fig:total_cpu_time}
\end{figure}

\begin{table}[!htp]
\centering
\scriptsize
\setlength{\tabcolsep}{2.7pt}
\caption{Execution time \textbf{in seconds} of the three FFS algorithms, KGroups (KG, on the left), KBest (KB, on the right), and mRMR (everything in between) per relevance estimator. The lowest execution time per relevance estimator is in \textbf{bold}. The last letter indicates the variant of mRMR: D for Difference and Q for Quotient. The other letters indicate the relevance and redundancy estimators e.g. RFCQ for RandomForest Pearson Correlation Quotient, MID for mutual information (for both relevance and redundancy) Difference, and FCQ for F-value Pearson Correlation Quotient.
}
\label{tab:per_dataset_cpu_time}
\begin{tabular}{@{}l|rrr|rrrr|rrrr@{}}
\toprule
& \multicolumn{3}{c|}{Gini import.} & \multicolumn{4}{c|}{Mutual info.} & \multicolumn{4}{c}{F-value} \\
\cmidrule(lr){2-4}\cmidrule(lr){5-8}\cmidrule(lr){9-12}
\textbf{Dataset} & \textbf{KG} & \textbf{RFCQ} & \textbf{KB} & \textbf{KG} & \textbf{MID} & \textbf{MIQ} & \textbf{KB} & \textbf{KG} & \textbf{FCD} & \textbf{FCQ} & \textbf{KB} \\
\midrule
ALLAML & \textbf{180} & 5,952 & 186 & \textbf{186} & 52,448 & 52,432 & 192 & 184 & 5,878 & 5,864 & \textbf{176} \\
CLL\_SUB\_111 & \textbf{234} & 9,600 & 275 & \textbf{253} & 94,996 & 94,983 & 271 & \textbf{233} & 9,434 & 9,425 & 245 \\
Carcinom & \textbf{552} & 8,307 & 917 & 752 & 96,504 & 94,879 & \textbf{746} & \textbf{28} & 8,004 & 7,998 & 660 \\
GLIOMA & 196 & 3,757 & \textbf{188} & \textbf{204} & 31,009 & 30,729 & 206 & \textbf{189} & 3,695 & 3,691 & 193 \\
GLI\_85 & \textbf{193} & 18,635 & 239 & \textbf{204} & 168,077 & 167,806 & 237 & 203 & 18,261 & 18,257 & \textbf{191} \\
Prostate & \textbf{194} & 5,136 & 205 & \textbf{202} & 49,347 & 48,072 & 209 & \textbf{195} & 5,050 & 5,045 & 196 \\
SMK\_CAN\_187 & \textbf{259} & 17,323 & 378 & \textbf{277} & 209,722 & 199,394 & 308 & 273 & 16,945 & 17,019 & \textbf{260} \\
TOX\_171 & \textbf{281} & 5,163 & 343 & \textbf{309} & 58,694 & 56,592 & 320 & \textbf{290} & 4,962 & 4,958 & 300 \\
colon & \textbf{181} & 1,860 & 174 & 190 & 14,564 & 14,644 & \textbf{185} & 187 & 1,794 & 1,790 & \textbf{184} \\
leukemia & 176 & 5,945 & \textbf{172} & \textbf{178} & 59,081 & 55,889 & 183 & 176 & 5,849 & 5,853 & \textbf{168} \\
lung & \textbf{276} & 3,287 & 324 & 311 & 36,602 & 37,106 & \textbf{307} & \textbf{270} & 3,096 & 3,086 & 290 \\
lung\_disc & \textbf{208} & 612 & 191 & 216 & 2,505 & 2,828 & \textbf{205} & 214 & 538 & 534 & \textbf{204} \\
lymphoma & 231 & 3,517 & \textbf{221} & \textbf{239} & 33,646 & 33,130 & 251 & 235 & 3,424 & 3,436 & \textbf{226} \\
nci9 & \textbf{216} & 7,869 & 217 & \textbf{232} & 81,428 & 76,530 & 274 & \textbf{178} & 7,849 & 7,778 & 209 \\
\bottomrule
\end{tabular}
\end{table}

\section{KGroups Limitations and Potentials}\label{sec:KGroups Limitations and Potentials}

In this section, we are going to discuss the limitations and potentials of the KGroups algorithm.

\subsection{KGroups Limitations}\label{subsec:KGroups Limitations}
Based on initial exploratory analysis of the distribution of the feature relevance values (independently of the relevance estimator), KGroups works better when the range is relatively large and the feature relevance values are fairly distributed across the range (not too many empty groups or clusters). For instance, the F-value relevance estimation of the nci9 dataset yielded only two non-empty clusters for $k\in[2,100]$. As a result, KGroups manages to select no more than two features during the low-dimension feature extraction task (see Fig. \ref{fig:num_selected_for_Fvalue}) and performed poorly (see Tables \ref{tab:Best overall configs - KGvsKB}-\ref{tab:Average best configs - KGvsMRMR}). 

Further study will be required to determine the exact or a more accurate estimation of the limits. However, for now, we can state that KGroups has two main limitations: the range and distribution of the feature relevance values. The second limitation seems to be the most important. One way to address it is to simply change the relevance estimator.
\subsection{KGroups Potentials}\label{subsec:KGroups Potentials}

Despite the limitations discussed in the above section, KGroups has two major advantages: speed and parameterisation; this makes KGroups suitable for hyperparameter finetuning. From a limited brute force hyperparameter finetuning performed in this study, we demonstrated that KGroups predictive performance can be improved from that procedure. Therefore, KGroups potentials can be divided into two: predictive performance optimisation and research opportunities.

\begin{figure}[!h]
\centering
\includegraphics[width=0.9\textwidth, height=8cm]{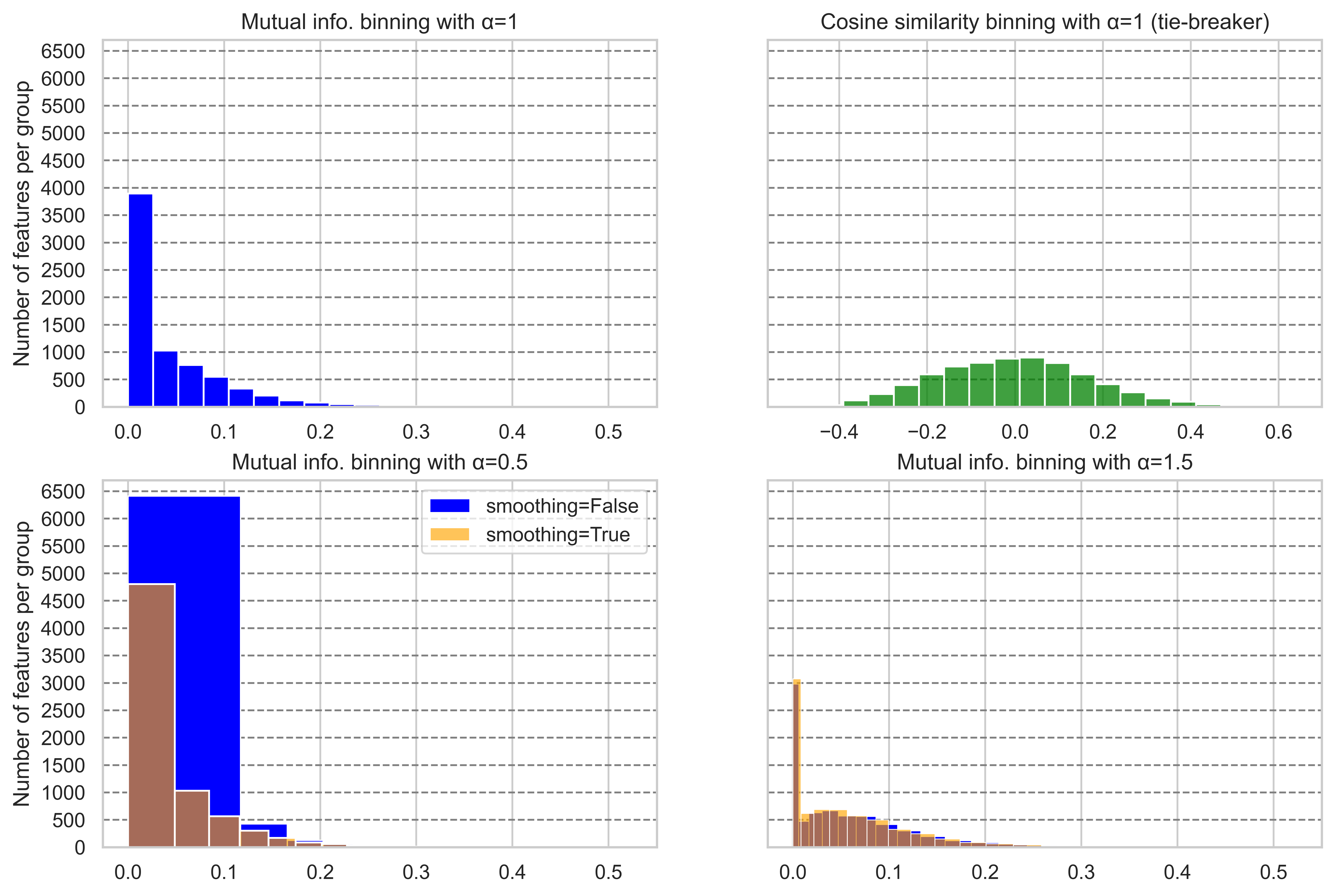}
\caption{Impact of $\alpha$ and the smoothing variable on the bins size for $k=20$; example of mutual information relevance estimator on the ALLAML dataset.}\label{fig:binning_resizing_and_smoothing}
\end{figure}

\textbf{Predictive performance optimisation.} If predictive performance is the main concern, then one can do a more complete brute force hyperparameter finetuning for $\alpha\in ]0; 3]$ with a step of 0.1 or 0.05, for example. One would have to train 30 and 60 variants of KGroups, respectively. Instead of a fixed step value of 0.1 or 0.05, one can use dynamic steps with larger values when close to 1 and smaller values when close to 0 or above 2. One can also finetune \textbf{the bin size smoothing hyperparameter, a boolean variable}. This boolean variable has two effects: (a) reduces the bins size of low relevance values for very small $\alpha$ and increase the bins size of low relevance values for large $\alpha$ (see Fig. \ref{fig:binning_resizing_and_smoothing}) and (b) converts fixed step into dynamic step values. This will raise the number of variants of KGroups to 60 and 120, respectively. This whole process is still computationally cheaper than one variant of mRMR; as aforementioned one variants of mRMR is equivalent to  123 variants of KGroups in terms of time complexity. KGroups predictive performance can also be improve further by trying a handful of tie-breakers for additional computational cost.

\textbf{Research opportunities.} Instead of a brute force approach of hyperparameter finetuning, researchers can investigate more efficient ways to achieve competing predictive performance with less computation. For instance, the choice of the tie-breaker(s) and the value(s) of $\alpha$ according to the distribution of the relevance estimator. Finding smarter and cheaper ways of making those two choices can make a huge difference. 

Figure \ref{fig:binning_resizing_and_smoothing} below shows the impact of $\alpha$ and the smoothing variable on the bins size for $k=20$. Mutual information relevance estimator on the ALLAML dataset is used as example. On the top row is displayed mutual information ($\alpha=1$, on the left) and cosine similarity (the tie-breaker, on the right) relevance values. One can notice that the distribution  mutual information relevance values is right-skewed (but with no empty clusters); however, the distribution of the cosine similarity relevance values is normally distributed. For $\alpha=1$, all the bins of the histogram have the same size. On the bottom row is shown the distribution of smoothed and non-smoothed bins for $\alpha=0.5$ and $\alpha=1.5$, respectively, from left to right. \textbf{For $\alpha < 1$, more features with high relevance values are selected. Inversely, more features with low relevance values are selected for $\alpha>1$.}

\section{Conclusion}\label{sec:Conclusion}
In this paper, we proposed KGroups, a new versatile univariate Max-Relevance Min-Redundancy algorithm that can easily be applied to any type of relevance estimator. KGroups is fast (up to 821 times faster that multivariate mRMR; 123 times faster on average) and parameterisable. This make it suitable for hyperparameters finetuning. Based on intensive and systematic experimentations, we demonstrated KGroups superiority compared to KBest. we also showed that a full hyperparameters finetuning of mRMR is equivalent to a limited brute force hyperparameters finetuning of KGroups in terms of predictive performance measured in average 5-fold cross validation accuracy. Next, we highlighted and discussed KGroups limitations and potentials. 

For future work, we aim to investigate ways to address KGroups limitations and harness its potentials. We also want to extend the experiments performed in this paper to additional datasets, especially to other types or modalities of data, such as images.

\backmatter

\bmhead{Supplementary information} The entire project, from the code files to the raw and processed experimental results, is released and publicly available on GitHub at \url{https://github.com/malick-jaures/KGroups_Bio.git} for ease of reproduction and to stimulate further research in this research area. 

\bmhead{Acknowledgements} This research was conducted with the financial support of Taighde E\'ireann – Research Ireland under Grant Agreement No. 13/RC/2106\_P2 at the ADAPT Centre at University College Dublin.

\bibliography{bibliography}

\end{document}